\pdfoutput=1

\documentclass[11pt]{article}

\usepackage{acl}

\usepackage{times}
\usepackage{latexsym}

\usepackage[T1]{fontenc}

\usepackage[utf8]{inputenc}

\usepackage{microtype}

\usepackage{booktabs}
\usepackage{amsmath} 
\usepackage{amssymb}
\usepackage{pifont}
\usepackage{diagbox}
\usepackage{array}
\usepackage{graphics}
\usepackage{multirow}
\usepackage{hyperref}

\usepackage{placeins} 

\usepackage{graphicx}
\newcommand{\CenterSmall}[1]{
\begin{center}
    \small#1
\end{center}
}

\title{Vision-and-Language Navigation: \\
A Survey of Tasks, Methods, and Future Directions}

\author{
Jing Gu$^1$\textsuperscript \,\,\, Eliana Stefani$^1$  \,\,\, Qi Wu$^2$  \,\,\, Jesse Thomason$^3$  \,\,\, Xin Eric Wang$^1$ \\
$^1$University of California, Santa Cruz \\
$^2$The University of Adelaide \quad $^3$University of Southern California\\
\texttt{\small \{jgu110,estefani,xwang366\}@ucsc.edu} \\
\texttt{\small qi.wu01@adelaide.edu.au, jessetho@usc.edu}
}

\begin{document}
\maketitle

\newcommand{\cmark}{\ding{51}}%
\newcommand{\xmark}{\ding{55}}%

\begin{abstract}
A long-term goal of AI research is to build intelligent agents that can communicate with humans in natural language, perceive the environment, and perform real-world tasks. 
Vision-and-Language Navigation (VLN) is a fundamental and interdisciplinary research topic towards this goal, and receives increasing attention from natural language processing, computer vision, robotics, and machine learning communities. 
In this paper, we review contemporary studies in the emerging field of VLN, covering tasks, evaluation metrics, methods, etc. 
Through structured analysis of current progress and challenges, we highlight the limitations of current VLN and opportunities for future work. This paper serves as a thorough reference for the VLN research community.\footnote{We also release a Github repo to keep track of advances in VLN: \url{https://github.com/eric-ai-lab/awesome-vision-language-navigation}}




\end{abstract}

\section{Introduction}
Humans communicate with each other using natural language to issue tasks and request help. An agent that can understand human language and navigate intelligently would significantly benefit human society, both personally and professionally. Such an agent can be spoken to in natural language, and would autonomously execute tasks such as household chores indoors, repetitive delivery work outdoors, or work in hazardous conditions following human commands (bridge inspection; fire-fighting). Scientifically, developing such an agent explores how an artificial agent interprets natural language from humans, perceives its visual environment, and utilizes that information to navigate to complete a task successfully.

\begin{figure}[t!]
    \centering
    \includegraphics[scale=0.26]{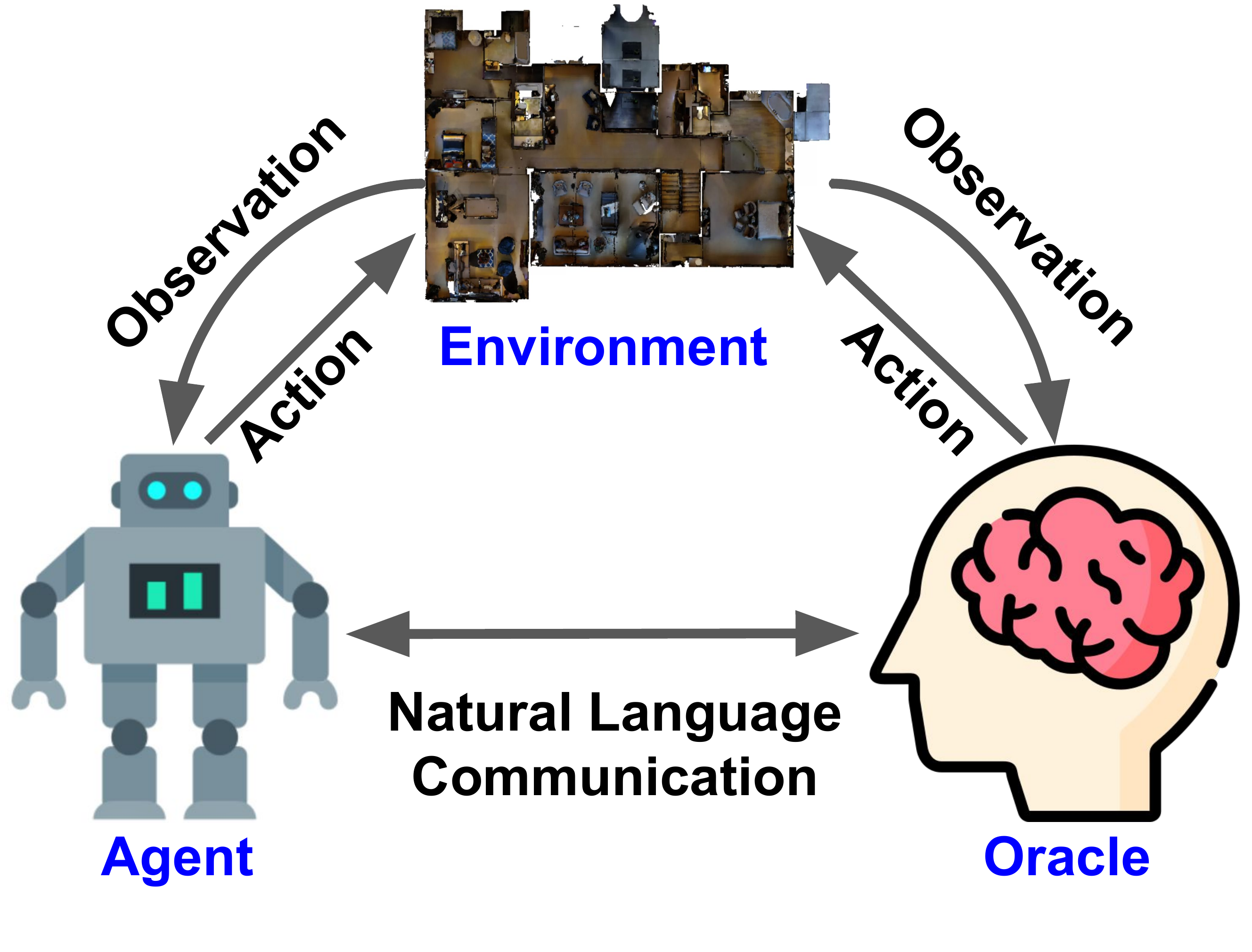}
    \vspace{-3ex}
    \caption{
    The agent and oracle discuss the VLN task in natural language. Both observe and interact with the navigable environment to accomplish a task.
    }
    \label{fig:relation}
\end{figure}

Vision-and-Language Navigation (VLN)~\citep{r2r, touchdown, thomason:corl19} is an emerging research field that aims to build such an embodied agent that can communicate with humans in natural language and navigate in real 3D environments. VLN extends visual navigation in both simulated~\citep{zhu2017target, 10.1145/3347450.3357659} and real environments~\citep{10.5555/3327144.3327168} with natural language communication.
As illustrated in Figure \ref{fig:relation}, VLN is a task that involves the oracle (frequently a human), the agent, and the environment.
The agent and the oracle communicate in natural language. The agent may ask for guidance and the oracle could respond. The agent navigates and interacts with the environment to complete the task according to the instructions received and the environment observed. Meanwhile, the oracle observes the environment and agent status, and may interact with the environment to help the agent.

Since the development and release of works such as Room-to-Room (R2R)~\citep{r2r},
many VLN datasets have been introduced.
Regarding the degree of communication, researchers create benchmarks where the agent is required to passively understand one instruction before navigation, to benchmarks where agents converse with the oracle in free-form dialog. Regarding the task objective, the requirements for the agent range from strictly following the route described in the initial instruction to actively exploring the environment and interacting with objects.
In a slight abuse of terminology, we refer to benchmarks that involve object interaction together with substantial sub-problems of navigation and localization, such as ALFRED~\cite{alfred}, as VLN benchmarks. 

Many challenges exist in VLN tasks. First, VLN faces a complex environment and requires effective understanding and alignment of information from different modalities. Second, VLN agents require a reasoning strategy for the navigation process. Data scarcity is also an obstacle. Lastly, the generalization of a model trained in seen environments to unseen environments is also essential. 
We categorize the solutions according to the respective challenges. (1) \emph{Representation learning} methods help understand information from different modalities. (2) \emph{Action strategy learning} aims to make reasonable decisions based on gathered information. (3) \emph{Data-centric learning} methods effectively utilize the data and address data challenges such as data scarcity. (4) \emph{Prior exploration} helps the model familiarize itself with the test environment, improving its ability to generalize.

We make three primary contributions. (1) We systematically categorize current VLN benchmarks from \emph{communication complexity} and \emph{task objective} perspectives, with each category focusing on a different type of VLN task. (2) We hierarchically classify current solutions and the papers within the scope. (3) We discuss potential opportunities and identify future directions.  




\section{Tasks and Datasets}
\label{sec:datasets}
    
    

\begin{table*}[ht]  
    \centering
    \small
    \begin{tabular}{p{5em}p{16em}p{12em}p{12em}}
    \toprule
    \multirow{2}{5em}{\bf Comm Complexity} & \multicolumn{3}{c}{\bf Task Objective} \\
    \cmidrule(lr){2-4}
    & Fine-grained Navigation & Coarse-grained Navigation & Nav + Object Interaction \\
    \midrule
    \begin{center} Initial Instruction(s) \end{center} & \small{Room-to-Room~\citep{r2r}, Room-for-Room~\citep{jain-etal-2019-stay}, Room-Across-Room~\citep{rxr},  XL-R2R~\citep{yan2020crosslingual},
    Landmark-RxR~\citep{landmark2021he},
    VLNCE~\citep{vlnce},
    TOUCHDOWN~\citep{touchdown},
    StreetLearn~\citep{mirowski2019streetlearn}, StreetNav~\citep{hermann2020learning},  Talk2Nav~\citep{vasudevan2021talk2nav}, LANI~\citep{misra2018mapping}} &  \CenterSmall{RoomNav~\citep{wu2018building}, EmbodiedQA~\citep{das2018embodied}, REVERIE~\citep{reverie}, SOON~\citep{zhu2021soon}} & \CenterSmall{IQA~\citep{gordon2018iqa}, CHAI~\citep{misra2018mapping}, ALFRED~\citep{alfred}}   \\ \midrule
    
    \begin{center}Oracle Guidance\end{center} & \CenterSmall{Just Ask~\citep{chi2020just}} &\small{VNLA~\citep{vnla}, HANNA~\citep{nguyen-daume-iii-2019-help}, CEREALBAR~\citep{suhr-etal-2019-executing}} & \CenterSmall{None}\\ \midrule
    
    \begin{center}Dialogue\end{center} & \CenterSmall{None}& \small{CVDN~\citep{thomason:corl19}, RobotSlang~\citep{banerjee:corl20}, Talk the Walk~\citep{talk_the_walk}} & \small{TEACh~\citep{padmakumar2021teach}, Minecraft Collaborative Building~\citep{narayan-chen-etal-2019-collaborative}, DialFRED~\citep{gao2022dialfred}} \\
    
    \bottomrule
    \end{tabular}
    \vspace{-1ex}
    \caption{Vision-and-Language Navigation benchmarks organized by \textbf{Communication Complexity} versus \textbf{Task Objective}. Please refer to Appendix for more details about the datasets and the commonly used underlying simulators. 
    }
    \label{tab:datasets}
\end{table*}

    
    

The ability for an agent to interpret natural language instructions (and in some instances, request feedback during navigation) is what makes VLN unique from visual navigation~\citep{bonin2008visual}. In Table \ref{sec:datasets}, we mainly categorize current datasets on two axes, \emph{Communication Complexity} and \emph{Task Objective}.

\textbf{Communication Complexity} defines the level at which the agent may converse with the oracle, and we differentiate three levels:
In the first level, the agent is only required to understand an \emph{Initial Instruction} before navigation starts. In the second level, the agent sends a signal for help whenever it is unsure, utilizing the \emph{Guidance} from the oracle. In the third level, the agent with \emph{Dialogue} ability asks questions in the form of natural language during the navigation and understands further oracle guidance.\par

\textbf{Task Objective} defines \textit{how} the agent attains its goal based on the \textit{initial instructions} from the oracle.
In the first objective type, \emph{Fine-grained Navigation}, the agent can find the target according to a detailed step-by-step route description. In the second type, \emph{Coarse-grained Navigation}, the agent is required to find a distant target goal with a coarse navigation description, requiring the agent to reason a path in a navigable environment and possibly elicit additional oracle help. Tasks in the previous two types only require the agent to navigate to complete the mission. In the third type, \emph{Navigation and Object Interaction}, besides reasoning a path, the agent also needs to interact with objects in the environment to achieve the goal since the object might be hidden or need to change physical states.\footnote{Navigation and Object Interaction includes both fine-grained and coarse-grained instructions, which ideally should be split further. But given that there are only few datasets in this category, we keep the current categorization in Table~\ref{sec:datasets}.}
As with coarse-grained navigation, some object interaction tasks can require additional supervision via dialogue with the oracle.


\subsection{Initial Instruction}
In many VLN benchmarks, the agent is given a natural language instruction for the whole navigation process, such as \textit{``Go upstairs and pass the table in the living room. Turn left and go through the door in the middle.''}

\noindent\textbf{Fine-grained Navigation~}~An agent needs to strictly follow the natural language instruction to reach the target goal. 
\citet{r2r} create the R2R dataset based on the Matterport3D simulator~\citep{Matterport3D}. An embodied agent in R2R moves through a house in the simulator traversing edges on a navigation graph, jumping to adjacent nodes containing panoramic views.
R2R is extended to create other VLN benchmarks. Room-for-Room joins paths in R2R to longer trajectories~\citep{jain-etal-2019-stay}. \citet{yan2020crosslingual} collect XL-R2R to extend R2R with Chinese instructions. 
RxR~\citep{rxr} contains instructions from English, Hindi, and Telegu. 
The dataset has more samples and the instructions in it are time-aligned to the virtual poses of the instruction. 
The English split of RxR is further extended to build Landmark-RxR~\citep{landmark2021he} by incorporating landmark information.



In most current datasets, agents traverse a navigation graph at predefined viewpoints. 
To facilitate transfer learning to real agents, VLN tasks should provide a continuous action space and a freely navigable environment. 
To this end, \citet{vlnce} reconstruct the navigation graph based R2R trajectories in continuous environments and create VLNCE. 
\citet{irshad2021hierarchical} propose Robo-VLN task where the agent operates in a continuous action space over long-horizon trajectories. 

Outdoor environments are usually more complex and contain more objects than indoor environments. In TOUCHDOWN~\citep{touchdown}, an agent follows instructions to navigate a streetview rendered simulation of New York City to find a hidden object. Most photo-realistic outdoor VLN datasets including TOUCHDOWN~\citep{touchdown}, StreetLearn~\citep{mirowski2019streetlearn, mehta-etal-2020-retouchdown}, StreetNav\citep{hermann2020learning}, and Talk2Nav~\citep{vasudevan2021talk2nav} are proposed based on Google Street View.

Some work uses natural language to guide drones. LANI~\citep{misra2018mapping} is a 3D synthetic navigation environment, where an agent navigates between landmarks following natural language instructions.  Current datasets on drone navigation usually fall in a synthetic environment such as Unity3D~\citep{blukis2018following, blukis2019learning}.


\noindent\textbf{Coarse-grained Navigation~}~In real life, detailed information about the route may not be available since it may be unknown to the human instructor (oracle). Usually, instructions are more concise and contain merely information of the target goal.



RoomNav~\cite{wu2018building} requires agent navigate according to instruction ``\textit{go to X}'', where X is a predefined room or object.

In Embodied QA~\citep{das2018embodied}, the agent navigates through the environment to find answer for a given question.
The instructions in REVERIE~\citep{reverie} are annotated by humans, and thus more complicated and diverse.
The agent navigates through the rooms and differentiates the object against multiple competing candidates. 
In SOON~\citep{zhu2021soon}, an agent receives a long, complex coarse-to-fine instruction which gradually narrows down the search scope.

\noindent\textbf{Navigation+Object Interaction~}~For some tasks, the target object might be hidden (e.g., the spoon in a drawer), or need to change status (e.g., a sliced apple is requested but only a whole apple is available). In these scenarios, it is necessary to interact with the objects to accomplish the task (e.g., opening the drawer or cutting the apple). 
Interactive Question Answering (IQA) requires the agent to navigate and sometimes to interact with objects to answer a given question.
Based on indoor scenes in AI2-THOR~\citep{ai2thor}, \citet{alfred} propose the ALFRED dataset, where agents are provided with both coarse-grained and fine-grained instructions complete household tasks in an interactive visual environment. CHAI~\citep{misra2018mapping} requires the agent to navigate and simply interact with the environments.

\subsection{Oracle Guidance}

Agents in Guidance VLN tasks may receive further natural language guidance from the oracle during navigation. For example, if the agent is unsure of the next step (e.g., entering the kitchen), it can send a [help] signal, and the oracle would assist by responding \textit{``go left''}~\citep{vnla}.

\noindent\textbf{Fine-grained Navigation~}~The initial fine-grained navigation instruction may still be ambiguous in a complex environment. Guidance from the oracle could clarify possible confusion. \citet{chi2020just} introduce Just Ask---a task where an agent could ask oracle for help during navigation. 

\noindent\textbf{Coarse-grained Navigation~}~With only a coarse-grained instruction given at the beginning, the agent tends to be more confused and spends more time exploring. Further guidance resolves this ambiguity.
VNLA~\citep{vnla} and HANNA~\citep{nguyen-daume-iii-2019-help} both train an agent to navigate indoors to find objects. The agent could request help from the oracle, which responds by providing a subtask which helps the agent make progress. While oracle in VNLA uses predefined script to respond, the oracle in HANNA uses a neural network to generate natural language responses. CEREALBAR~\citep{suhr-etal-2019-executing} is a collaborative task between a leader and a follower. Both agents move in a virtual game environment to collect valid sets of cards.

\noindent\textbf{Navigation+Object Interaction~} While VLN is still in its youth, there are no VLN datasets in support of Guidance and Object Interaction.


\subsection{Human Dialogue}

It is human-friendly to use natural language to request help~\citep{banerjee:corl20, thomason:corl19}. For example, when the agent is not sure about what fruit the human wants, it could ask \textit{``What fruit do you want, the banana in the refrigerator or the apple on the table?''}, and the human response would provide clear navigation direction.


\noindent\textbf{Fine-grained Navigation~}~No datasets are in the scope of this category. Currently, route-detailed instruction with possible guidance could help the agent achieve relatively good performance in most simulated environments. We expect datasets to be developed for this category for super long horizon navigation tasks in complex environments especially with rich dynamics where dialog is necessary to clear confusions.

\noindent\textbf{Coarse-grained Navigation~}~CVDN~\citep{thomason:corl19} is a dataset of human-human dialogues. 
Besides interpreting a natural language instruction and deciding on the following action, the VLN agent also needs to ask questions in natural language for guidance. The oracle, with knowledge of the best next steps, needs to understand and correctly answer said questions. 

Dialogue is important in complex outdoor environments. \citet{talk_the_walk} introduce the Talk the Walk dataset, where the guide has knowledge from a map and guides the tourist to a destination, but does not know the tourist's location; while the tourist navigates a 2D grid via discrete actions.

\noindent\textbf{Navigation+Object Interaction~}~Minecraft Collaborative Building~\citep{narayan-chen-etal-2019-collaborative} studies how an agent places blocks into a building by communicating with the oracle.
TEACh~\citep{padmakumar2021teach} is a dataset that studies object interaction and navigation with free-form dialog. The follower converses with the commander and interacts with the environment to complete various house tasks such as making coffee. DialFRED~\citep{gao2022dialfred} extends ALFRED~\citep{alfred} dataset by allowing the agent to actively ask questions.

\section{Evaluation}
\label{sec:evaluation}



\noindent\textbf{Goal-oriented Metrics~}~ mainly consider the agent's proximity to the goal.
The most intuitive is \emph{Success Rate (SR)}, which measures how frequently an agent completes the task within a certain distance of the goal. \emph{Goal Progress}~\citep{thomason:corl19} measures the reduction in remaining distance to the target goal.
\emph{Path Length (PL)} measures the total length of the navigation path.
\emph{Shortest-Path Distance (SPD)} measures the mean distance between the agent’s final location and the goal.
Since a longer path length is undesirable (increases duration and wear-and-tear on actual robots), \emph{Success weighted by Path Length (SPL)} ~\citep{anderson2018evaluation} balances both Success Rate and Path Length. Similarly, \emph{Success weighted by Edit Distance (SED)}~\citep{touchdown} compares the expert's actions/trajectory to the agent's actions/trajectory, also balancing SR and PL.
\emph{Oracle Navigation Error (ONE)} takes the shortest distance from any node in the path rather than just the last node, and \emph{Oracle Success Rate (OSR)} measures whether any node in the path is within a threshold from the target location.


\noindent\textbf{Path-fidelity Metrics~} evaluate to what extent an agent follows the desired path. Some tasks require the agent not only to find the goal location but also to follow specific path.
Fidelity measures the matches between the action sequence in the expert demonstration and the action sequence in the agent trajectory. 
\emph{Coverage weighted by LS (CLS)}~\citep{jain-etal-2019-stay} is the product of the \emph{Path Coverage (PC)} and \emph{Length Score (LS)} with respect to the reference path. It measures how closely an agent’s trajectory follows the reference path.
\emph{Normalized Dynamic Time Warping (nDTW)}~\citep{ilharco2019general} softly penalizes deviations from the reference path to calculate the match between two paths.
\emph{Success weighted by normalized Dynamic Time Warping (SDTW)}~\citep{ilharco2019general} further constrains nDTW to only successful episodes to capture both success and fidelity.

\section{VLN Methods}
\label{sec:methods}

\begin{figure}[t!]
    \centering
    \includegraphics[scale=0.25]{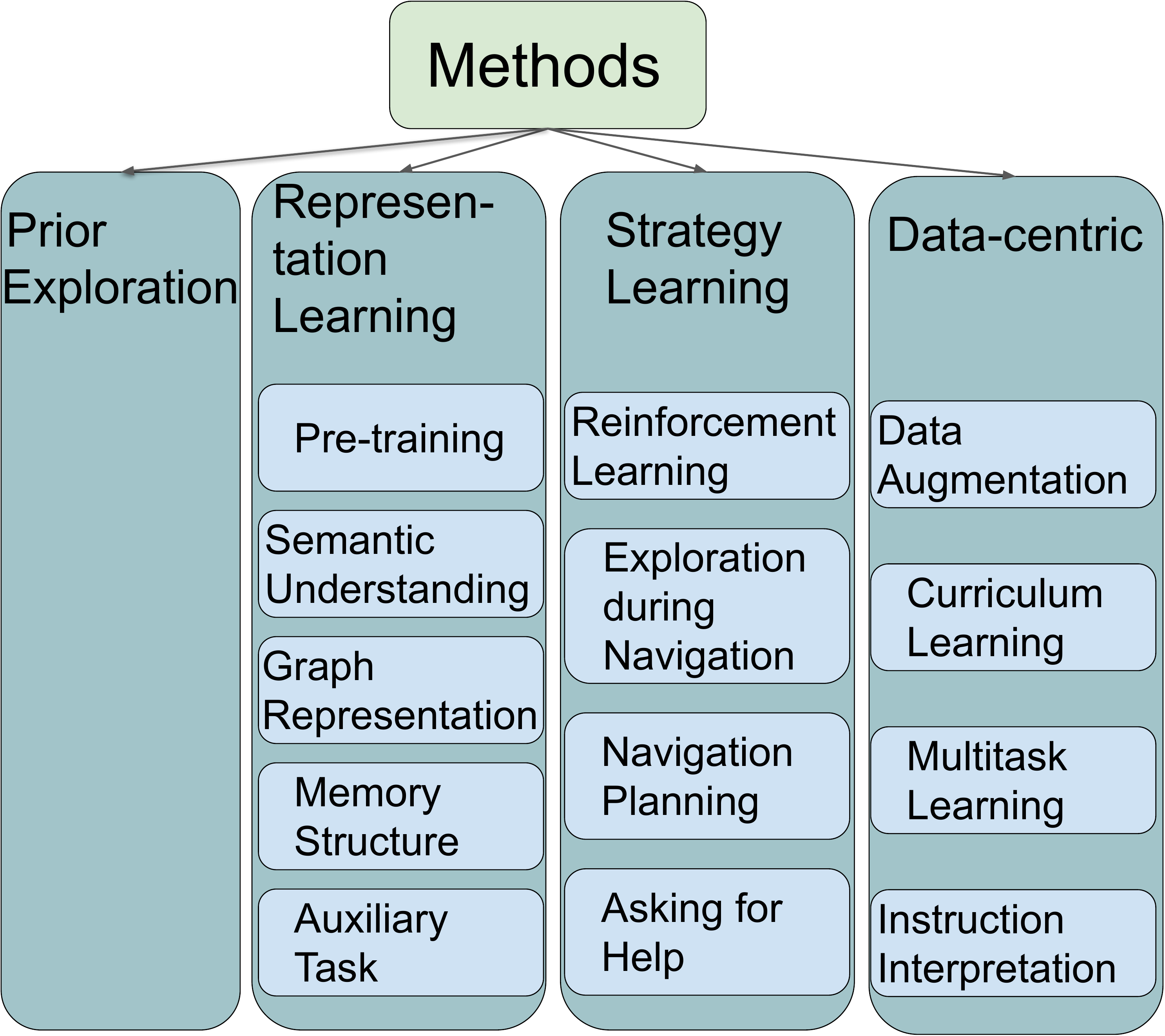}
    \vspace{-1ex}
    \caption{Categories of VLN methods. Methods may not be mutually exclusive to an individual category.}
    \label{fig:methods}
\end{figure}

As shown in Figure~\ref{fig:methods}, we categorize existing methods into \emph{Representation Learning}, \emph{Action Strategy Learning}, \emph{Data-centric Learning}, and \emph{Prior Exploration}. 
Representation learning methods help agent understand relations between these modalities since VLN involves multiple modalities, including vision, language, and action.
Moreover, VLN is a complex reasoning task where mission results depend on the accumulating steps, and better action strategies help the decision-making process.
Additionally, VLN tasks face challenges within their training data. One severe problem is scarcity. Collecting training data for VLN is expensive and time-consuming, and the existing VLN datasets are relatively small with respect to the complexity of VLN tasks. Therefore, data-centric methods help to utilize the existing data and create more training data.
Prior exploration helps adapt agents to previously unseen environments, improving their ability to generalize, decreasing the performance gap between seen versus unseen environments.




\subsection{Representation Learning}

Representation learning helps the agent understand how the words in the instruction relate to the perceived features in the environment.


\subsubsection{Pretraining}


\noindent\textbf{Vision or Language~}~Using a pretrained model to initialize a vision or text encoder provides agents with single-modality knowledge. pretrained vision models may use a ResNet~\citep{he2016deep} or Vision Transformers~\citep{dosovitskiy2020image}. Other navigation tasks~\citep{wijmans2019dd} may also provide visual initialization~\citep{vlnce}.
Large pretrained language models such as BERT~\citep{devlin2019bert} and GPT~\citep{radford2019language} can encode language and improve instruction understanding~\citep{li2019robust}, which can be further pretrained with VLN instructions ~\citep{Pashevich_2021_ICCV} before fine-tuning in VLN task.

\noindent\textbf{Vision and Language~}~Vision-and-language pretrained models provide good joint representation for text and vision. A common practice is to initialize a VLN agent~\citep{kim2021ndh} with a pretrained model such as ViLBERT~\citep{NEURIPS2019_c74d97b0}. The agent may be further trained with VLN-specific features such as objects and rooms~\citep{Qi_2021_ICCV}.


\noindent\textbf{VLN~}~Downstream tasks benefit from being closely related to the pretraining task. Researchers also explored pretraining on the VLN domain directly. VLN-BERT~\citep{majumdar2020improving} pretrains navigation models to measure the compatibility between paths and instructions, which formats VLN as a path selection problem.
PREVALENT \citep{hao2020prevalent} is trained from scratch on image-text-action triplets to learn textual representations in VLN tasks.
The output embedding from the [CLS] token in BERT-based pretraining models could be leveraged in a recurrent fashion to represent history state~\citep{VLN_bert, moudgil2021soat}.
Airbert~\citep{Guhur_2021_ICCV} achieve good performance on few-shot setting after pretraining on a large-scale in-domain dataset. 

\subsubsection{Semantic Understanding}



Semantic understanding of VLN tasks incorporates knowledge about important features in VLN. In addition to the raw features, high-level semantic representations also improve performance in unseen environments.

\noindent\textbf{Intra-Modality~}~Visual or textual modalities can be decomposed into many features, which matter differently in VLN.
The overall visual features extracted by a neural model may not be important and may actually hurt the performance in some cases~\citep{thomason-etal-2019-shifting, hu-etal-2019-looking, ijcai2020-0124, schumann2022analyzing}. Therefore, it is important to find the feature(s) that best improve performance.
High-level features such as visual appearance, route structure, and detected objects outperform the low level visual features extracted by CNN~\citep{hu-etal-2019-looking}. Different types of tokens within the instruction also function differently~\citep{zhu2021diagnosing}. Extracting these tokens and encoding the object tokens and directions tokens are crucial~\citep{qi2020object, zhu2021diagnosing}. 


\noindent\textbf{Inter-Modality~}~Semantic connections between different modalities: actions, scenes, observed objects, direction clues, and objects mentioned in instructions can be extracted and then softly aligned with attention mechanism~\citep{qi2020object, gao2021room}.
The soft alignment also highlights relevant parts of the instruction with respect to the current step~\citep{landi2019embodied, zhang2020language}.

\subsubsection{Graph Representation}

Building graph to incorporate structured information from instruction and environment observation provides explicit semantic relation to guide the navigation.
The graph neural network may encode the relation between text and vision to better interpret the context information~\citep{hong2020language, deng-2020-evolving}.
The graph could record the location information during the navigation, which can used to predict the most likely trajectory~\citep{vln-chasing-ghosts} or probability distribution over action space~\citep{deng-2020-evolving}. When connected with prior exploration, an overview graph about the navigable environment~\citep{chen2021topological} can be built to improve navigation interpretation.




\subsubsection{Memory-augmented Model}


Information accumulates as the agent navigates, which is not efficient to utilize directly. Memory structure helps the agent effectively leverage the navigation history. 
Some solutions leverage memory modules such as LSTMs or recurrently utilize informative states~\citep{VLN_bert}, which can be relatively easily implemented, but may struggle to remember features at the beginning of the path as path length increases.
Another solution is to build a separate memory model to store the relevant information~\citep{zhu2020vision, lin2021scene, nguyen-daume-iii-2019-help}. Notably, by hierarchically encoding a single view, a panorama, and then all panoramas in history, HAMT~\citep{chen2021history} successfully utilized the full navigation history for decision-making.




\subsubsection{Auxiliary Tasks}

Auxiliary tasks help the agent better understand the environment and its own status without extra labels. From the machine learning perspective, an auxiliary task is usually achieved in the form of an additional loss function. 
The auxiliary task could, for example, explain its previous actions, or predict information about future decisions~\citep{Zhu_2020_CVPR}.
Auxiliary tasks could also involve the current mission such as current task accomplishment, and vision \& instruction alignment~\citep{ma2019self, Zhu_2020_CVPR}.
Notably, auxiliary tasks are effective when adapting pretrained representations for VLN~\citep{Huang_2019_ICCV}.   







\subsection{Action Strategy Learning}


With many possible action choices and complicated environment, action strategy learning provides a variety of methods to help the agent decide on those best actions.

\subsubsection{Reinforcement Learning}

VLN is a sequential decision-making problem and can naturally be modeled as a Markov decision process. So Reinforcement Learning (RL) methods are proposed to learn better policy for VLN tasks.
A critical challenge for RL methods is that VLN agents only receive the success signal at the end of the episode, so it is difficult to know which actions to attribute success to, and which to penalize. 
To address the ill-posed feedback issue, \citet{cvpr2019wang, wang2020vision} propose RCM model to enforces cross-modal grounding both locally and globally, with goal-oriented extrinsic reward and instruction-fidelity intrinsic reward. \citet{landmark2021he} propose to utilize the local alignment between the instruction and critical landmarks as the reward. Evaluation metrics such as CLS~\citep{jain-etal-2019-stay} or nDTW~\citep{ilharco2019general} can also provide informative reward signal~\citep{landi2020perceive}, and natural language may also provide suggestions for reward~\citep{fu2019language}.

To model the dynamics in the environment, \citet{wang2018look}
leverage model-based reinforcement learning to predict the next state and improve the generalization in unseen environment. \citet{zhang2020language} find recursively alternating the learning schemes of imitation and reinforcement learning improve the performance.

\subsubsection{Exploration during Navigation}

Exploring and gathering environmental information while navigating provides a better understanding of the state space.
Student-forcing is a frequently used strategy, where the agent keeps navigating based on sampled actions and is supervised by the shortest-path action~\citep{r2r}. \par
There is a tradeoff between exploration versus exploitation: with more exploration, the agent sees better performance at the cost of a longer path and longer duration, so the model needs to determine when and how deep to explore ~\citep{wang2020active}. After having gathered the local information, the agent needs to decide which step to choose, or whether to backtrack~\citep{ke2019tactile}. Notably, \citet{Koh_2021_ICCV} designed Pathdreamer, a visual world model to synthesize visual observation future viewpoints without actually looking ahead.




\subsubsection{Navigation Planning}

Planing future navigation steps leads to a better action strategy. 
From the visual side, predicting the waypoints~\citep{Krantz_2021_ICCV}, next state and reward~\citep{wang2018look}, generate future observation~\citep{Koh_2021_ICCV} or incorporating neighbor views~\citep{an2021neighbor} has proven effective. Recognizing and stopping at the correct location also reduces navigation costs~\citep{xiang2020learning}.
The natural language instruction also contains landmarks and direction clues to plan detailed steps. \citet{anderson2019chasing} predict the forthcoming events based on the instruction, which is used to predict actions with a semantic spatial map. \citep{kurita2020generative} formulates VLN as a generative approach where a language model is used to compute the distribution over all possible instructions. The instruction may also be used to tag navigation and interaction milestones which the agent needs to complete step by step~\citep{raychaudhuri2021language, song2022one}.




\subsubsection{Asking for Help}

An intelligent agent asks for help when uncertain about the next action~\citep{https://doi.org/10.48550/arxiv.2110.08258}. 
Action probabilities or a separately trained model~\citep{chi2020just, Zhu_2021_ICCV, https://doi.org/10.48550/arxiv.2110.08258} can be leveraged to decide whether to ask for help.
Using natural language to converse with the oracle covers a wider problem scope than sending a signal. Both rule-based methods~\citep{padmakumar2021teach} and neural-based methods~\citep{roman:emnlpf20, https://doi.org/10.48550/arxiv.2110.08258} have been developed to build navigation agents with dialog ability.
Meanwhile, for tasks~\citep{thomason:corl19, padmakumar2021teach} that do not provide an oracle agent to answer question in natural language, researchers also need to build a rule-based~\citep{padmakumar2021teach} or neural-based~\citep{roman:emnlpf20} oracle. DialFRED~\citep{gao2022dialfred} uses a language model as an oracle to answer questions.

\subsection{Data-centric Learning}


Compared with previously discussed works that focus on building a better VLN agent structure, data-centric methods most effectively utilize the existing data, or create synthetic data. 

\subsubsection{Data Augmentation}




\noindent\textbf{Trajectory-Instruction Augmentation~}~Augmented path-instruction pairs could be used in VLN directly. Currently the common practice is to train a speaker module to generate instructions given a navigation path~\citep{fried2018speaker}. 
This generated data have varying quality~\citep{zhao2021evaluation, huang2019multimodal}. Therefore an alignment scorer~\citep{Huang_2019_ICCV} or adversarial discriminator~\citep{fu2020aps} can select high-quality pairs for augmentation. Style transfer module may also improve instruction quality via adapting instructions from the source domain~\citep{zhu2021multimodal}.


\noindent\textbf{Environment Augmentation~}~Generating more environment data not only helps generate more trajectories, but also alleviates the problem of overfitting in seen environments. Randomly masking the same visual feature across different viewpoints~\citep{tan-etal-2019-learning}, splitting the house scenes and re-mixing them~\citep{liu2021visionlanguage}, or changing style and objects~\citep{li2022envedit} could create new environments, which could further be used to generate more trajectory-instruction pairs~\citep{fried2018speaker}. Training data may also be augmented by replacing some visual features with counterfactual ones~\citep{parvaneh2020counterfactual}.



\subsubsection{Curriculum Learning}

Curriculum learning~\citep{bengio2009curriculum} gradually increases the task's difficulty during the training process.
The instruction length could be a metric for task difficulty. BabyWalk~\citep{zhu2020babywalk} keep increasing training samples' instruction length during the training process.
Attributes from the trajectory may also be used to rank task difficulty.
\citet{curriculum2021zhang} rearrange the R2R dataset using the number of rooms each path traverses. They found curriculum learning helps smooth the loss landscape and find a better local optima. 



\subsubsection{Multitask Learning}


Different VLN tasks can benefit from each other by cross-task knowledge transfer.
\citet{wang2020environmentagnostic} propose an environment-agnostic multitask navigation model for both VLN and Navigation from Dialog History tasks~\citep{thomason:corl19}.
\citet{ijcai2020-338} propose an attention module to train a multitask navigation agent to follow instructions and answer questions~\citep{eqa_matterport}.


%

\subsubsection{Instruction Interpretation}


A trajectory instruction interpreted multiple times in different ways may help the agent better understand its objective. LEO~\citep{xia2020multiview} leverages and encodes all the instructions with a shared set of parameters to enhance the textual understanding. LWIT~\citep{nguyen2021look} interprets the instructions to make it clear to interact with what class of objects.
Shorter, and more concise instructions provide clearer guidance for the agent compared to longer, semantically entangled instructions, thus \citet{hong-etal-2020-sub} breaks long instructions into shorter ones, allowing the agent to track progress and focus on each atomic instruction individually.




\subsection{Prior Exploration}

Good performance in seen environments often cannot generalize to unseen environments~\citep{hu-etal-2019-looking, parvaneh2020counterfactual, tan-etal-2019-learning}. Prior exploration methods allow the agent to observe and adapt to unseen environments,\footnote{Thus prior exploration methods are not directly comparable with other VLN methods.} bridging the performance gap between seen and unseen environments.


\citet{cvpr2019wang} introduce a self-supervised imitation learning to learn from the agent's own past, good behaviors. The best navigation path determined to align the instruction the best by a matching critic will be used to update the agent.
\citet{tan-etal-2019-learning} leverage the testing environments to sample and augment paths for adaptation.
\citet{fu2020aps} propose environment-based prior exploration, where the agent can only explore a particular environment where it is deployed.
When utilizing graph, prior exploration may construct a map or overview about the unseen environment to provide explicit guidance for navigation~\citep{chen2021topological, zhou2021rethinking}. 

\section{Related Visual-and-Language Tasks}
\label{sec:related}


This paper focuses on Vision-and-Language Navigation tasks with an emphasis on photo-realistic environments. 2D map may also be a uesful virtual environment for navigation tasks~\citep{vogel2010learning, chen2011learning, paz2019run}. Synthetic environments may also be a substitute for realistic environment~\citep{macmahon2006walk, pmlr-v100-blukis20a}.
\citet{tellex2011understanding} propose to instantiate a probabilistic graphical model for natural language commands in robotic navigation and mobile manipulation process.

In VLN, an agent needs to follow the given instruction and even ask for assistants in human language. An agent in Visual Navigation tasks is usually not required to understand information from textual modality. Visual Navigation~\citep{zhu2021deep} is a problem of navigating an agent from the current location to find the goal target. Researchers have achieved success in both simulated environments~\citep{zhu2017target, 10.1145/3347450.3357659} and real environments~\citep{10.5555/3327144.3327168}.



\section{Conclusion and Future Directions}
\label{sec:future-direction}


In this paper, we discuss the importance of VLN agents as a part of society, how their tasks vary as a function of communication level versus task objective, and how different agents may be evaluated. We broadly review VLN methodologies and categorize them. This paper only discusses these issues broadly at an introductory level. In reviewing these papers, we can see the immense progress that has already been made, as well as directions that this research topic can be expanded on. \par

Current methods usually do not explicitly utilize external knowledge such as objects and general house descriptions in Wikipedia. Incorporating knowledge also improves the interpretability and trust of embodied AI.
Moreover, currently several navigation agents learn which direction to move and with what to interact, but there is a last-mile problem of VLN---how to interact with objects. \citet{r2r} asked whether a robot could learn to \textit{``Bring me a spoon''}; new research may ask how a robot can learn to \textit{``Pick up a spoon''}.
The environments also lack diversity: most interior terrestrial VLN data consists of American houses, but never warehouses or hospitals: the places where these agents may be of most use.

Below we detail additional future directions:

\noindent\textbf{Collaborative VLN~}
Current VLN benchmarks and methods predominantly focus on tasks where only one agent navigates, yet complicated real-world scenarios may require several robots collaborating. Multi-agent VLN tasks require development in swarm intelligence, information communication, and performance evaluation. MeetUp!~\citep{ilinykh2019meetup} is a two-player coordination game where players move in a visual environment to find each other.
VLN studies the relationship between the human and the environment in Figure~\ref{fig:relation}, yet here humans are oracles simply observing (but not acting on) the environment. Collaboration between humans and robots is crucial for them to work together as teams (e.g., as personal assistants or helping in construction). Future work may target at collaborative VLN between multiple agents or between human and agents.

\noindent\textbf{Simulation to Reality~}
There is a performance loss when transferred to real-life robot navigation~\citep{vln-pano2real}.
Real robots function in continuous space, but most simulators only allow agents to ``hop'' through a pre-defined navigation graph which is unrealistic for three reasons~\cite{vlnce}. Navigation graphs assume: (1) perfect localization---in the real world it is a noisy estimate; (2) oracle navigation---real robots cannot ``teleport'' to a new node; (3) known topology---in reality an agent may not have access to a preset list of navigable nodes.
Continuous implementations of realistic environments may contain patches of the images, be blurred, or have parallax errors, making them unrealistic. A simulation that is based on both a 3D model and realistic imagery could improve the match between virtual sensors (in simulation) and real sensors.
Lastly, most simulators assume a static environment only changed by the agent. This does not account for other dynamics such as people walking or objects moving, nor does it account for lighting conditions through the day. VLN environments with probabilistic transition functions may also narrow the gap between simulation and reality.

\noindent\textbf{Ethics \& Privacy~}~
During both training and inference, VLN agents may observe and store sensitive information that can get leaked or misused. Effective navigation with privacy protection is crucially important. Relevant areas such as federated learning~\citep{konevcny2016federated} or differential privacy~\citep{dwork2006calibrating} could also be studied in VLN domain to preserve the privacy of training and inference environments.


\noindent\textbf{Multicultural VLN~}~
VLN lacks diversity in 3D environments: most outdoor VLN datasets use Google Street View recorded in major American cities, but lacks data in developing countries. Agents trained on American data face potential generalization problems in other city or housing layouts.
Future work should explore more diverse environments across multiple cultures and regions.
Multilingual VLN datasets~\citep{yan2020crosslingual,rxr} could be good resources to study multicultural differences from the linguistic perspective. 




\section*{Acknowledgement}

We thank anonymous reviewers, Juncheng Li, Yue Fan, Tongzhou Jiang for their feedback.

\bibliography{anthology,custom}
\bibliographystyle{acl_natbib}

\clearpage

\appendix
\section{Dataset Details}
\label{appendix-dataset-details}

Here in Table~\ref{tab:app_dataset_details}, we introduce more information about the datasets. Compared with the number of the datasets, the simulators are limited. More specifically, most indoor datasets are based on Matterport3D and most outdoor datasets are based on Google Street View. Also, more datasets are about indoor environments rather than outdoor environments. Outdoor environments are usually more complex and contain more objects compared with indoor environments.

\begin{table*}[!b]

    \centering
    \begin{tabular}{cccc}
        \toprule
          Name & Simulator & Language-Active & Environment \\
         \midrule
         
          Room-to-Room~\citep{r2r} & Matterport3D & \xmark  & Indoor\\
          Room-for-Room~\citep{jain-etal-2019-stay} & Matterport3D & \xmark  & Indoor\\
          Room-Across-Room~\citep{rxr} & Matterport3D  & \xmark  & Indoor\\
          Landmark-RxR~\citep{landmark2021he} & Matterport3D & \xmark  & Indoor\\
          XL-R2R~\citep{yan2020crosslingual} & Matterport3D & \xmark  & Indoor\\
          VLNCE~\citep{vlnce} & Habitat & \xmark  & Indoor\\
          StreetLearn~\citep{mirowski2019streetlearn} & Google Street View & \xmark & Outdoor\\
          StreetNav~\citep{hermann2020learning} & Google Street View & \xmark & Outdoor\\
          TOUCHDOWN~\citep{touchdown} & Google Street View & \xmark & Outdoor\\
          Talk2Nav~\citep{vasudevan2021talk2nav} & Google Street View & \xmark & Outdoor\\
          LANI~\citep{misra2018mapping} & - & \xmark& Outdoor \\

         RoomNav~\citep{wu2018building} & House3D & \xmark & Indoor\\
         EmbodiedQA~\citep{das2018embodied} & House3D & \xmark & Indoor\\
          REVERIE~\citep{reverie} & Matterport3D & \xmark & Indoor\\
          SOON~\citep{zhu2021soon} & Matterport3D & \xmark & Indoor\\

         IQA~\citep{gordon2018iqa} & AI2-THOR & \xmark & Indoor\\
         CHAI~\citep{misra2018mapping} & CHALET & \xmark & Indoor \\
         ALFRED~\citep{alfred} & AI2-THOR & \xmark & Indoor\\

         VNLA~\citep{vnla} & Matterport3D & \cmark & Indoor\\
         HANNA~\citep{nguyen-daume-iii-2019-help} & Matterport3D & \cmark & Indoor\\
         CEREALBAR~\citep{suhr-etal-2019-executing} & - & \cmark & Indoor\\
         Just Ask~\citep{chi2020just} & Matterport3D & \cmark & Indoor\\
         CVDN~\citep{thomason:corl19} & Matterport3D & \cmark & Indoor\\
         RobotSlang~\citep{banerjee:corl20} & - & \cmark & Indoor\\
         Talk the Walk~\citep{talk_the_walk} & - & \cmark & Outdoor\\
         MC Collab~\citep{narayan-chen-etal-2019-collaborative} & Minecraft & \cmark & Outdoor\\
         TEACh~\citep{padmakumar2021teach} & AI2-THOR & \cmark & Indoor \\
         DialFRED~\citep{gao2022dialfred} & AI2-THOR & \cmark & Indoor \\

         \bottomrule
    \end{tabular}
    \caption{Vision-and-Language Navigation datasets. Language-Active means the agent needs to use natural language to request help, including both Guidance datasets and Dialog datasets in Table~\ref{tab:datasets}.}
    \label{tab:app_dataset_details}
\end{table*}

\section{Simulator}
\label{appendix-simulator}

The virtual features of the dataset are deeply connected with the simulator in which datasets are built. Here we summarize simulators frequently used during the VLN dataset creation process.

House3D~\citep{wu2018building} is a realistic virtual 3D environment built based on the SUNCG~\citep{SUNCG} dataset. An agent in the environment has access to first-person view RGB images, together with semantic/instance masks and depth information.

Matterport3D~\citep{r2r} simulator is a large-scale visual reinforcement learning simulation environment for research on embodied AI based on the Matterport3D dataset~\citep{Matterport3D}. Matterport3D contains various indoor scenes, including houses, apartments, hotels, offices, and churches. An agent can navigate between viewpoints along a pre-defined graph. Most indoors VLN datasets such as R2R and its variants are based on the Matterport3D simulator.

Habitat~\citep{habitat19iccv, szot2021habitat} is a 3D simulation platform for training embodied AI in 3D physics-enabled scenarios. Compared with other simulation environments, Habitat 2.0~\citep{szot2021habitat} shows strength in system response speed. Habitat has the following datasets built-in: Matterport3D~\citep{Matterport3D}, Gibson~\citep{Xia_2018_CVPR}, and Replica~\citep{replica19arxiv}. 
AI2-THOR~\citep{ai2thor} is a near photo-realistic 3D indoor simulation environment, where agents could navigate and interact with objects. Based on the object interaction function, it helps to build a dataset that requires object interaction, such as ALFRED~\citep{alfred}.

Gibson~\citep{Xia_2018_CVPR} is a real-world perception interactive environment with complex semantics.  Each viewpoint has a set of RGB panoramas with global camera poses and reconstructed 3D meshes. Matterport3D dataset~\citep{Matterport3D} is also integrated into the Gibson simulator.

House3D~\citep{wu2018building} converts SUNCG’s static environment into a virtual environment, where the agent can navigate with physical constraints (e.g.
it cannot pass through walls or objects).

LANI~\citep{misra2018mapping} is a 3D simulator built in Unity3D platform. The environment in LANI is a fenced, square, grass field containing randomly placed landmarks. An agent needs to navigate between landmarks following the natural language instruction. Drone navigation tasks~\citep{blukis2018following, blukis2019learning} are also built based on LANI.

Currently, most datasets and simulators focus on indoors navigable scenes partly because of the difficulty of building an outdoor photo-realistic 3D simulator out of the increased complexity.
Google Street View~\footnote{\url{https://developers.google.com/maps/documentation/streetview/overview}}, an online API that is integrated with Google Maps, is composed of billions of realistic street-level panoramas. It has been frequently used to create outdoor VLN tasks since the development of TOUCHDOWN~\citep{touchdown}.

\begin{table*}[!hb]
    \centering
    \begin{tabular}{ccc}
        \toprule
		Simulator & Photo-realistic & 3D \\ \toprule
		House3D~\citep{wu2018building} & \cmark  & \cmark \\ 
		Matterport3D~\citep{Matterport3D} & \cmark & \cmark \\
		Habitat~\citep{habitat19iccv} & \cmark & \cmark \\ 
		AI2-THOR~\citep{ai2thor}  & \xmark  & \cmark \\ 
		Gibson~\citep{Xia_2018_CVPR} & \cmark  & \cmark \\ 
		LANI~\citep{misra2018mapping} & \xmark & \cmark \\
		*Google Street View & \cmark & \cmark \\
		\bottomrule
    \end{tabular}
    \caption{Common simulators used to build VLN datasets. *Google Street View is online API, providing similar functionality as a simulator for building VLN datasets.}
    \label{tab:my_label}
\end{table*}
\section{Room-to-Room Leaderboard}
\label{appendix-r2r-leaderboard}

\begin{table*}[p]
	\small
	\begin{center}
	    \scalebox{0.7}{

				\begin{tabular}{l | c c  c c c | c c c c c| c c c c c}
					\hline
					Leader-Board (Test Unseen) 
					& \multicolumn{5}{c|}{Single Run} 
					& \multicolumn{5}{c|}{Prior Exploration}
					& \multicolumn{5}{c}{Beam Search}\\
					\hline
					Models
					& TL$\downarrow$ & NE$\downarrow$ & OSR$\uparrow$ & SR$\uparrow$& SPL$\uparrow$
					& TL$\downarrow$ & NE$\downarrow$ & OSR$\uparrow$ & SR$\uparrow$& SPL$\uparrow$
					& TL$\downarrow$ & NE$\downarrow$ & OSR$\uparrow$ & SR$\uparrow$& SPL$\uparrow$ \\
					\hline
					Random & 9.89 &  9.79 &  0.18 &  0.13 & 0.12  & - & - & - & - & - & - & - & - & - & -\\
					Human & 11.85 & 1.61 & 0.90 & 0.86 & 0.76  & - & - & - & - & - & - & - &&&\\
					\hline
					Seq-to-Seq~\citep{r2r} & 8.13 &  20.4 &  0.27 &  0.20 & 0.18  & - & - & - & - & - & - & - &-&-&-\\
					RPA~\citep{wang2018look} & 9.15 &  7.53 &   0.32 & 0.25  & 0.23  &-   &-  &- &- & - & - & - & - & - & -\\
					Speaker-Follower~\citep{fried2018speaker} & 14.82 & 6.62 & 0.44  & 0.35  & 0.28  & -  & -  & - &- & -& 1257.38 & 4.87 & 0.96 & 0.54 & 0.01 \\
					 Chasing Ghosts~\citep{vln-chasing-ghosts} & 10.03 & 7.83 &  0.42 & 0.33  & 0.30  &  - & - &- &- & - & - & - &- &- & -\\
                     Self-Monitoring~\citep{ma2019self}& 18.04 & 5.67 & 0.59 & 0.48 & 0.35 & -  & - & -& -&  -& 373.1 & 4.48 & 0.97 & 0.61 & 0.02  \\
                     RCM~!\citep{cvpr2019wang}&11.97 & 6.12 & 0.50 & 0.43 & 0.38 & 9.48	& 4.21 & 0.67 & 0.60 & 0.59 &  357.6 & 4.03 & 0.96 & 0.63 & 0.02  \\
					 Regretful Agent~\citep{ma2019theregretful}&  13.69 & 5.69 & 0.56 & 0.48 & 0.40  & -  & - & -& -& - & - &- & - & - &- \\
                     FAST~\citep{ke2019tactile} & 22.08 & 5.14 & 0.64 & 0.54 & 0.41 &- & - & -& -& - & 196.5 & 4.29 & 0.90 & 0.61 & 0.03 \\
                     ALTR~\citep{Huang_2019_ICCV} & 10.27 & 5.49 & 0.56 & 0.48 & 0.45  & -  & - & -& -& - & - & & - & - &- \\
					 EnvDrop~\citep{tan-etal-2019-learning} & 11.66 & 5.23 & 0.59 & 0.51 & 0.47 & 9.79 & 3.97 & 0.70 & 0.64 & 0.61 & 686.8 & 3.26 & 0.99 & 0.69 & 0.01 \\
					 PRESS~\citep{li2019robust} & 10.52 & 4.53 &0.63 & 0.57 & 0.53  &   -& - &- &- &-  &-  &-& - & - & - \\
					 PTA~\citep{landi2020perceive}& 10.17 & 6.17 & 0.47 & 0.40 & 0.36  & -  & - & -& -& - & - & -\\
					 EGP~\citep{deng-2020-evolving} & - &  5.34 & 0.61 & 0.53 & 0.42  &  - & - & -& -& - & - &- & - & - &-\\
                     SERL~\citep{Wang-2020-Soft} & 12.13 & 5.63 & 0.61 & 0.53 & 0.49  & -  &-  & -& -& - & 690.61 & 3.21 & 0.99 & 0.70 & 0.01\\
					 OAAM~\citep{qi2020object} & 10.40 &  - & 0.61 & 0.53 & 0.50  &  - & - &- & -& - & - &- & - & - &-\\
                     CMG-AAL~\citep{zhang2020language} & 12.07 & 3.41 & 0.76 & 0.67 & 0.60  & -  & - & -& -& - &  &- & - & - &- \\
					 AuxRN~\citep{Zhu_2020_CVPR} & - & 5.15 & 0.62 & 0.55 & 0.51 & 10.43 & 3.69 & 0.75 & 0.68 & 0.65 & 40.85 & 3.24 & 0.81 & 0.71 & 0.21 \\
					 RelGraph~\citep{hong2020language} & 10.29 & 4.75 & 0.61 & 0.55 & 0.52 & - & - &- &- & - &-  & - & - & - & - \\
					 PRRVALENT~\citep{hao2020prevalent}&10.51 & 5.30 & 0.61 & 0.54 & 0.51  & - & - & -& -& - &-&- & - & - & -\\
					 Active Exploration~\citep{wang2020active}& 21.03 & 4.34 & 0.71 & 0.60 & 0.43 & 9.85 & 3.30 & 0.77 & 0.70 & 0.68 & 176.2 & 3.07 & 0.94 & 0.70 & 0.05\\
					 VLN-BERT~\citep{majumdar2020improving}  & -& -& -& -& -& -& -& -& -& -& 686.62 & 3.09 & 0.99 & 0.73 & 0.01 \\
                     DASA~\citep{9428422} & 10.06 & 5.11 & - & 0.54 & 0.52  & -  & - & -& -& - & - & -& - & - & -\\
					 ORIST~\citep{Qi_2021_ICCV} & 11.31 & 5.10 & - & 0.57 & 0.52  & -  & -  & - & -& -& - & - &- & - & -\\
					 NvEM~\citep{an2021neighbor} & 12.98 & 4.37 & 0.66 & 0.58 & 0.54  & -  &-  &- & -& - & - &- & - & - & -\\
					 SSM~\citep{Wang_2021_CVPR} & 20.39 & 4.57 & 0.70 & 0.61 & 0.46 & -  &-  &- & -& - &-  &-& - & - & - \\
					 Recurrent VLN BERT~\citep{VLN_bert}& 12.35	& 4.09 & 0.70 & 0.63 & 0.57 &  - & - & -& -& - & - &- & - & - & -\\
					 SOAT~\citep{moudgil2021soat} & 12.26 & - & 4.49 & 58 & 53 & &&&&&&&& & \\
					 REM~\citep{liu2021visionlanguage}& 13.11 & 3.87 & 0.72 & 0.65 & 0.59  & -  & - & -& -& - & - & -& - & - & -\\
					 HAMT\citep{chen2021history} & 12.27 & 3.93 & 0.72 & 0.65 & 0.60 &-&-&-&-&-&-&-&-&-&-  \\
					 Spatial Route Prior~\citep{zhou2021rethinking} & -& -& -& -& -& -& -& -& -& -& 625.27 & 3.55 & 0.99 & 0.74 & 0.01 \\
					 Airbert~\citep{Guhur_2021_ICCV} & -& -& -& -& -& -& -& -& -& -& 686.54 & 2.58 & 0.99 & 0.78 & 0.01	\\ 
					 3DSR~\citep{tan2022self} & 15.89 & 3.73 & 0.73 & 0.66 & 0.60 &-&-&-&-&-&-&-&-&-&-\\
					\hline
					
				\end{tabular}
		}
	\end{center}
	\caption{Leaderboard of Room-to-Room benchmark as of March, 2022}
	\label{table:r2r-leaderboard}
\end{table*}

Room-to-Room (R2R)~\citep{r2r} is the benchmark used most frequently for evaluating different methods. Here we collect all the reported performance metrics in the corresponding papers and the official R2R leaderboard\footnote{\url{https://eval.ai/web/challenges/challenge-page/97/leaderboard/270}}. Since beam search explores more routes, and since prior exploration has additional observations in the test environment, their performance can not be directly compared with other methods.  

\end{document}